\documentclass[11pt]{article}
\usepackage{acl}
\usepackage{times,latexsym}
\usepackage[T1]{fontenc}
\usepackage{subcaption}

\usepackage{times}
\usepackage{latexsym}
\usepackage[utf8]{inputenc}

\usepackage[authormarkup=none]{changes}

\usepackage{inconsolata}
\usepackage{amsmath}
\usepackage{amsthm}
\usepackage{amsfonts}
\usepackage{amssymb}
\usepackage{booktabs}
\usepackage{mathtools}
\usepackage{bbm}
\usepackage{graphicx}
\usepackage{microtype}
\usepackage{fontawesome}

\def\vcontext{\mathbf{w}_{<t}}
\def\vclass{\mathcal{C}_i}
\def\vword{w_{t}}
\def\vvocab{\mathcal{W}}

\usepackage{tikz}
\usetikzlibrary{bayesnet}
\usetikzlibrary{arrows}
\usetikzlibrary{decorations}

\usepackage{graphicx}
\usepackage{booktabs}
\usepackage{tabularx}
\usepackage{tabulary}
\usepackage{colortbl}

\usepackage{xspace}
\usepackage{xcolor}
\newcolumntype{Y}{>{\centering\arraybackslash}X}
  \newcolumntype{P}{>{\raggedleft\arraybackslash}X}
\usepackage{multirow}
\usepackage{stfloats}
\usepackage{diagbox}

\usepackage{algorithm}
\usepackage[noend]{algorithmic}

\usepackage[noabbrev,capitalise]{cleveref}
\usepackage{pgffor}
\newcommand{\seq}{\,{=}\,}
\newcommand{\slt}{\,{<}\,}
\definechangesauthor[name=After reviews, color=brown]{AR}

\usepackage{amsmath,amsfonts,bm}

\def\eqref#1{equation~\ref{#1}}

\def\1{\bm{1}}

\DeclareMathAlphabet{\mathsfit}{\encodingdefault}{\sfdefault}{m}{sl}
\SetMathAlphabet{\mathsfit}{bold}{\encodingdefault}{\sfdefault}{bx}{n}

\usepackage{todonotes}
\makeatletter
\newcommand*\iftodonotes{\if@todonotes@disabled\expandafter\@secondoftwo\else\expandafter\@firstoftwo\fi}
\makeatother

\definecolor{edolime}{rgb}{0.9,1,0.3}

\definecolor{colemanorange}{rgb}{0.9,0.25,0}

\title{A Grounded Typology of Word Classes}

\author{Coleman Haley\textsuperscript{\faFont}~~~Sharon Goldwater\textsuperscript{\faFont}~~~Edoardo Ponti\textsuperscript{\faFont \faImage}\\
\textsuperscript{\faFont}University of Edinburgh~~~\textsuperscript{\faImage}University of Cambridge\\
\texttt{\{coleman.haley,sgwater,eponti\}@ed.ac.uk}
}

\date{}

\begin{document}
\maketitle

\begin{abstract}

We propose a grounded approach to meaning in language typology. We treat data from perceptual modalities, such as images, as a language-agnostic representation of meaning. Hence, we can quantify the function--form relationship between images and captions across languages. Inspired by information theory, we define ``groundedness'', an empirical measure of contextual semantic contentfulness (formulated as a difference in surprisal) which can be computed with multilingual multimodal language models. As a proof of concept, we apply this measure to the typology of word classes. Our measure captures the contentfulness asymmetry between functional (grammatical) and lexical (content) classes across languages, but contradicts the view that functional classes do not convey content. Moreover, we find universal trends in the hierarchy of groundedness (e.g., nouns > adjectives > verbs), and show that our measure partly correlates with psycholinguistic concreteness norms in English. We release a dataset of groundedness scores for 30 languages. Our results suggest that the grounded typology approach can provide quantitative evidence about semantic function in language.
\end{abstract}
\section{Introduction}

Within linguistics, {\em typology} is the subfield focused on the study of patterns and variation across the world's languages \citep[pp. 1--2]{croft-typology-2002}. To identify such patterns, linguists must carefully identify phenomena of interest within languages, and then align them with one another.
 For example, vowels exist in a continuous acoustic and perceptual space, without clear boundaries between them. To define vowel categories and align systems across languages, linguists rely largely on acoustic properties of the speech signal---reducing the problem to a physically grounded, empirical one \citep{liljencrants-numerical-1972, cotterell-eisner-2017-probabilistic}. 

While empirically grounding language form (surface structure like vowels) is typically straightforward, language is not just a formal system, but also a functional one. Many questions within typology relate to the relationship between form and {\em meaning}, especially in domains like morphology and syntax. Typically, typologists manually identify semantic/functional roles such as ``subject'', and ``causative'' and study their expression across languages \citep{haspelmath-comparative-2010, greenberg-universals-1966}. Unlike with many definitions based on form, definitions based on meaning are left up to subjective discretion, leading to debates which reduce to the definition of particular terms cross-linguistically \citep{haspelmath-preestablished-2007, haspelmath-how-2012, plank-inflection-1994}. %

\begin{figure}
  \centering
  \includegraphics[width=0.85\linewidth]{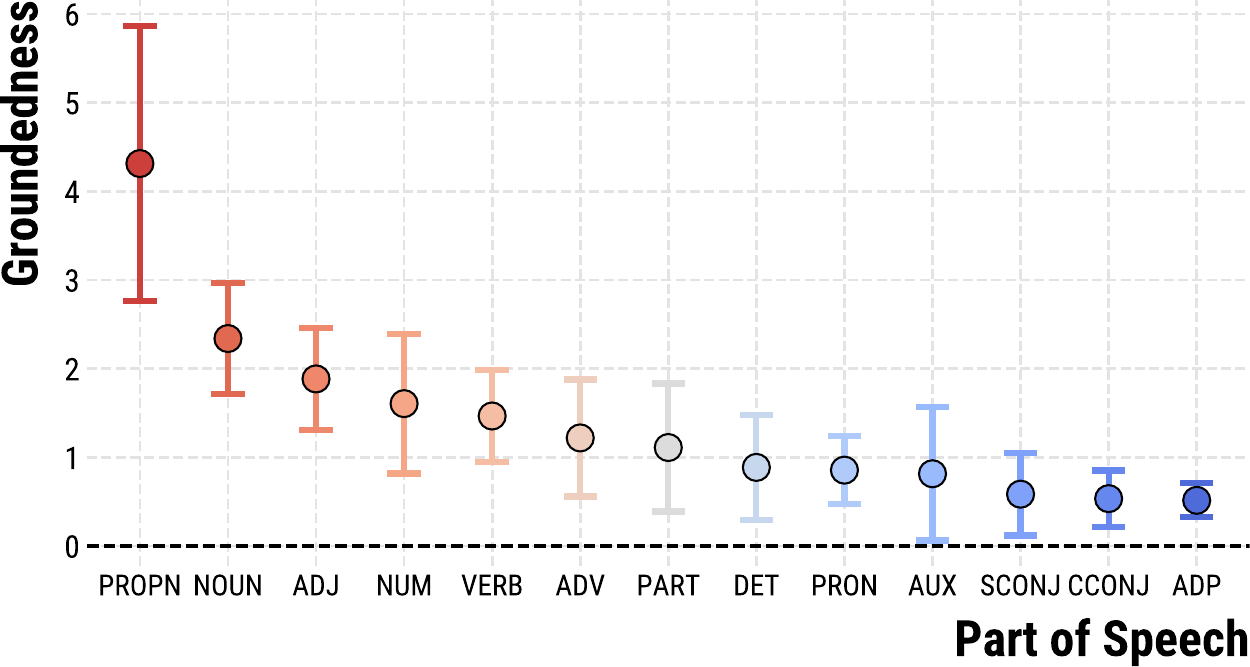}
  \caption{Mean and standard deviation of per-language mutual information estimates between word class and image. Across 30 languages, we see clear and consistent tendencies about which parts of speech are more ``grounded'', corresponding to a distinction between lexical and functional classes.}
  \label{fig:avg}
\end{figure}

Instead, we propose a ``grounded'' approach to typology, which (under certain assumptions), allows the quantification and cross-linguistic comparison of language function and semantics across languages. By looking at sentences produced as captions of the same image across languages, we can use the image as an evidence-based, language-agnostic representation of the shared semantics underlying these utterances, similar to the evidence-based acoustic signal in the study of vowel spaces.

In this work, we specifically focus on semantic contentfulness---how semantically informative a given word token is. We introduce a way to empirically quantify contentfulness, {\em groundedness}, which relies on vision-and-language models. Groundedness measures how much less surprising a word is when we know the perceptual stimuli (i.e., the image) it describes. This \textit{surprisal difference} between the surprisal of the word token in an image captioning model versus its surprisal in a language model is an estimate of the pointwise mutual information: the greater this difference (LM > captioning), the more \textit{grounded} the word is in that context.

As a case study, we apply this measure to the study of the typology of word classes (``parts of speech''). Literature from cognitive, pyscho- and neurolinguistics all point to contentfulness being an organizing factor in word class processing and even formation and structure: low-content (functional) word classes have many different properties from high-content (lexical) classes \citep{dube-independent-2014, bird-verbs-2003, chiarello-imageability-1999}. Yet, there has been no cross-linguistic study of the relationship between contentfulness and word class.

Using our groundedness measure to quantify semantic contentfulness, we can estimate the mutual information of a word class with a caption's meaning (image). We find our measure largely rediscovers the distinction between lexical and functional word classes across 30 languages. Further, though it correlates only weakly with psycholinguistic norms for imageability and concreteness in English, it provides an intuitive ranking (noun > adjectives > verbs) across languages. On the other hand, it contradicts the view of adpositions as a ``semi-lexical'' class \citep{corver-semilexical-2001} and suggests grammatical word classes do carry some semantic content. These results thus partly validate and partly falsify received wisdom about word class contentfulness. They suggest the utility of this measure as a general tool for studying contentfulness in linguistics, and of taking a grounded approach to typological problems. We release the model used to estimate our measure and a dataset of groundedness values in 30 languages.\footnote{\url{https://osf.io/bdhna/}}

\section{Background}%

An excellent example of the relevance of the relationship between semantic function and linguistic form to typology is {\em word classes}. Within a particular language, there are typically groups of words unified by the (formal) contexts in which they can appear. Further, this distribution of words is not arbitrary, but unified by a particular semantic prototype. For example, in English, nouns are a class of words which prototypically denote physical objects or things and can follow words like ``\textit{the}'', ``\textit{this}'', and ``\textit{that}''. However, not all languages have words like ``\textit{the}'', and so an equivalent formal--structural criterion cannot be given \citep{haspelmath-how-2012}. On the other hand, semantic criteria are not sufficient to describe these classes: most languages can express prototypical verb or adjective meanings with the syntactic distribution of a noun.

The elusiveness of a cross-linguistic definition for word classes leads to many debates about particular languages ``having'' or ``not having'' a distinction between (e.g.) nouns and verbs on the basis of a mix of formal and semantic criteria \citep[cf.][]{kaufman-austronesian-2009, hsieh-distinguishing-2019, richards-nouns-2009, weber-grammar-1983, floyd-rediscovering-2011}.
Here, we investigate word classes as operationalized in a  framework where there is a fixed set of {\em universally applicable} word classes, as set out in the Universal Dependencies project \citep{demarneffe-universal-2021}.%
While this is problematic in general, our aim is not to claim that the assignment of word classes is precisely correct, but rather to empirically and quantitatively investigate the functional/semantic dimension of this common operationalisation of word class. In future work, we aim to investigate the relationship between these measures and non-prototypical parts of speech.

\subsection{Contentfulness and word class}
In this work, we focus on the related distinction between lexical/contentful word classes (e.g. nouns, verbs, and adjectives) and functional/grammatical word classes. Functional word classes are typically closed-class, meaning they do not admit new members and typically do not exhibit rich productive morphology; they tend to express highly grammatical and abstract meanings. Lexical classes are typically open class, productively admitting new members, and their meanings tend to be more concrete and contentful \citep{corver-semilexical-2001}. 

Complications about these generalized categories and tendencies abound, however. For example, in some languages like Jaminjung, prototypically lexical categories like verbs are closed class \citep{schultze-berndt-simple-2000, pawley2006where}. Further, both the abstraction and semantic contentfulness of particular members of a given word class can be quite variable. For example, a noun like ``\textit{factor}'' has a highly abstract meaning, while the meaning of the preposition ``\textit{to}'' is intuitively more abstract than the preposition ``\textit{above}'', despite belonging to the same, ``abstract'' grammatical word class. Further, over time words can change in both their contentfulness and even word class through processes like grammaticalization \citep{bisang-grammaticalization-2017}.

Nevertheless, the complex relationship between contentfulness and word class remains unexplored through a cross-linguistic empirical lens---perhaps due to the difficulties of measuring such properties.

\subsection{Measuring contentfulness}
The relationship between contentfulness and word class has not been explored cross-linguistically; however, a significant literature within the language sciences has investigated related concepts.%

While theoretical linguistics has focused on a distinction between content and function words, psycholinguistics has focused on semantic dimensions like  imageability, concreteness, and strength of perceptual experience. Measures of these dimensions have relied on subjective, decontextualized human judgements, 
but nevertheless predict processing differences between word classes, such as asymmetries in the processing of nouns and verbs in certain aphasias \citep{bird-verbs-2003, dube-independent-2014, lin-word-2022}. Because we operationalize meaning as images, notions such as imageability seem especially related to our groundedness measure. However, as discussed in Section~\ref{sec:human}, these concepts  differ from our measure in that informativity is not a major factor in their definition. For example, while both ``zebra'' and ``woman'' are highly concrete nouns, the former has higher groundedness on average, because 
although both are often strongly associated with an image, ``zebra'' is more informative/surprising, especially if the image is unavailable---thus, the image adds more information in that case.

As shown by the prior example, our measure is also closely related to another concept widely studied in computational psycholinguistics: {\em surprisal.} Like our groundedness measure, surprisal has an intuitive link to contentfulness from an information theoretic perspective, and has been extensively studied in relation to processing difficulty \citep{hale-2001-probabilistic,levy-2008-expectation,smith-levy-2013,wilcox-etal-2023-testing,staub-predictability-2024a}. However, surprisal entangles formal and functional information in language. As such, cross-linguistic comparisons based on surprisal are challenging, since form is language specific \citep{park-etal-2021-morphology}. We aim to focus on information due to language {\em function}, separated from form. Surprisal must also encode grammatical uncertainty (alternative ways of expressing the same meaning like ``knight'' and ``cavalier''), as opposed to surprisal due only to what meanings are being expressed. Our image captioning model quantifies how many bits of information remain after the meaning is known. Our measure then quantifies how much of the LM surprisal is explained by the meaning (image).

\section{Method}%
In this section, we define a token's {\em groundedness}, and show how we can use this to estimate the mutual information between parts of speech and representations of meaning.
Let the set of word types in a language be $\vvocab$. We assume a model of the data generation process where given a meaning $m$, a sentence is constructed by iteratively sampling a word $w_t\in\vvocab$ conditioned on $m$ and previous words $\mathbf{w}_{<t}$. As mentioned previously, the groundedness of a token is given by its pointwise mutual information (PMI) with the meaning.
\begin{align} \label{pmi}
        \text{PMI}(w_t; m \mid \vcontext) = \log \frac{p(w_t \mid m, \mathbf{w}_{<t})}{p(w_t \mid \mathbf{w}_{<t})}
\end{align}
As we cannot access the true meaning $m$, we must approximate it with a proxy. A good proxy for $m$ should be language-neutral, and will make estimating the probabilities in Equation~\ref{pmi} straightforward across languages. In this work, we focus on {\em images} as a language-neutral representation of meaning. Images capture rich, language-independent information about the world state described by an image, and have proved useful as a method for aligning meanings across languages \citep{rajendran-etal-2016-bridge, gella-etal-2017-image, mohammadshahi-etal-2019-aligning, wu-etal-2022-leveraging}. Further, a major strength of images as a meaning representation is that estimating both quantities in Equation~\ref{pmi} becomes straightforward with neural models: $p_{\bm\phi}(w_t\hspace{-0.25em}\mid\hspace{-0.25em}m, \mathbf{w}_{<t})$ corresponds to the probability of the token under an image captioning model, while $p_{\bm\theta}(w_t \hspace{-0.25em}\mid\hspace{-0.25em}\mathbf{w}_{<t})$ corresponds to its probability under a language model.%

Using images as a representation of meaning does have some implications for our approach. For instance, verbs, which usually denote events and are more temporally unstable \citep{givon1984syntax} than other parts of speech, may be less grounded than with a different meaning representation, such as videos. Further, the language of image captions is somewhat restricted in terms of grammatical structure and lexical items, making the analysis of long-tail phenomena or highly abstract language challenging \citep{ferraro-etal-2015-survey,alikhani-stone-2019-caption}.
Future work could use our framework to explore other meaning representations, such as symbolic models or videos (though doing so involves overcoming further dataset and modeling challenges).
Still, the language-neutral nature and rich information content of images allows us to study groundedness for a wide range of words, languages, and linguistic contexts.

Noting that a model's surprisal is negative log probability,  we can view groundedness as a {\em difference in surprisal}, corresponding to how much more expected the token is under the grounded model than under the textual model. As such,
the PMI should rarely take on negative values---because the captioning model has more information (both image and text) than the language model (text only). However, some tokens, such as those that are highly grammatical or structural, should be close to 0. %

In this work, we study the groundedness of {\em word classes}. Drawing inspiration from functionalist typology, we treat a word class $\vclass$ as a label selected by a linguist for a word in its context. We make an assumption that this label is independent of our meaning representation given a word's context, allowing us to define the following joint distribution:
\begin{align}
    &p(\vclass, m \mid \vcontext) =  \nonumber\\ 
    &\quad\;\sum_{\vword\in\vvocab}\big[ p(\vclass \mid \vword,\vcontext) p(w_t, m \mid \vcontext) \big].
\end{align}
We can then formulate the mutual information between a word class and meaning as the expected value of the PMI between each token labeled with that class, and the token's associated image:
\begin{align}
I[\vclass; m | \vcontext]
    &= \mathbb{E}_{\mathclap{\hspace{-0.5em}\raisebox{-0.5em}{\scalebox{0.6}{${p(\vclass, m, \vcontext)}$}}}}\hspace{1.1em} \left[ \log \frac{ p(\vword | \vcontext, m)}{ p(\vword | \vcontext))} \right]. 
\end{align}
Given our factorization of the joint, we can perform a Monte Carlo estimation of the expectation by simply averaging groundedness over all the tokens tagged with $\vclass$ in the data $\mathcal{D}$:
\begin{align} \label{mihat}
        &\hat{I}[\mathcal{C}_i; m \mid \vcontext] = \nonumber \\
        &\quad\;\sum_{\mathclap{(m, \mathbf{w}_{<t}) \in \mathcal{D}}} \frac{\mathbbm{1}_{\mathcal{C}_{\vword}=\vclass}  \log \frac{p_{\bm\phi}(w_t \mid \mathbf{w}_{<t}, m)}{p_{\bm\theta}(w_t \mid \mathbf{w}_{<t})}}{\sum_{w_t \in \mathcal{D}} \mathbbm{1}_{\mathcal{C}_{\vword}=\vclass} }
\end{align}
where $\mathbbm{1}_{\mathcal{C}_{\vword}=\vclass}$ is 1 when a token's class is $C_i$ and 0 otherwise. We note that our groundedness measure and our mutual information estimates are conditional on {\em linguistic context}. As such, words which are very grounded in one context could be hardly grounded in another, due to disambiguating information in the preceding context. Some information about $m$ will be generally conveyed by $\vcontext$; however, our mutual information estimates are aggregated over all contexts in which a word class occurs, and on average this contribution is small.

\section{Experimental setup}%
\begin{table}[t]
    \centering
    {\small
    \begin{tabularx}{0.9\linewidth}{r | c c c}
    \toprule
    \multirow{2}{*}{\textbf{Model}} & Gemma & PaliGemma & COCO-35L \\
    & PT & CT & FT \\
    \midrule
    Img. Cap. & \faFont & \faImage\,\faFont & \faImage\,\faFont \\
    LM & \faFont & \faImage\,\faFont & \faFont \\
    \bottomrule
    \end{tabularx}
    }
    \caption{We match the data points on which the language model and image captioning model were trained. The three datasets are the Gemma pre-training mixture (PT), PaliGemma multimodal data for continued training (CT), and COCO image--caption pairs for fine-tuning (FT). Symbols indicate whether models are trained on text data (\faFont) or on multimodal data (\faImage\,\faFont).}
    \label{tab:datasets}
\end{table}

\paragraph{Captioning model $p_{\bm\phi}(w_t\hspace{-0.25em}\mid\hspace{-0.25em}\mathbf{w}_{<t}, m)$} As our image captioning model, we use the recently released PaliGemma model \citep{beyer2024paligemmaversatile3bvlm}. This model is by far the state-of-the-art among publicly available multilingual image captioning models.
PaliGemma consists of an image encoder, initialized from the SigLIP-So400m model \citep{zhai-sigmoid-2023}, and a transformer decoder language model, initialized from the Gemma-2B language model \citep{gemmateam2024gemmaopenmodelsbased}. A linear projection maps from the image encoder space to a sequence of 256 tokens in the language model's embedding space. The whole system is then trained on a mix of vision-and-language datasets, including the unreleased WebLI dataset with 10 billion image-caption pairs in 109 languages \citep{chen-pali-2023}, and the CC3M-35L dataset consisting of 3 million image-caption pairs in each of 35 languages \citep{thapliyal-crossmodal3600-2022}.

While PaliGemma is a general-purpose vision-and-language model, %
it is designed to be fine-tuned on and applied to individual tasks. As such, we use the open-source \texttt{paligemma-3b-ft-coco35-224} checkpoint for multilingual captioning, which has been fine-tuned on COCO-35L.

\paragraph{Language model $p_{\bm\theta}(w_t\hspace{-0.25em}\mid\hspace{-0.25em}\mathbf{w}_{<t})$} Our aim is to use a language model as similar to our captioning model $p_{\bm\phi}(w_t\hspace{-0.25em}\mid\hspace{-0.25em}\mathbf{w}_{<t}, m)$ as possible. This is critical to getting good (P)MI estimates, which relies on estimating a difference in surprisal between the two models. If the language model is not adapted to the image captioning domain, it may under-estimate the probability of particular words, leading to an over-estimation of mutual information. We therefore aim to {\em match} the training data between the language model and image captioning model, such that they see the same set of captions. 

To do so, we initialize our language model with the weights from the pretrained  PaliGemma model \texttt{paligemma-3b-pt-224}. However, out of the box, 
the decoder behaves degenerately when no image is provided, so we need to adapt the model to not expect image information and to match the training data of the captioning model. To do so, we fine-tune the language model on the {\em captions only} from the COCO-35L dataset. In this way, we ensure the models have observed the same data during training and are adapted to the same domain, and are thus maximally comparable. Table~\ref{tab:datasets} summarizes the data matching between the two models. Further implementational and POS tagging details are in Appendix~\ref{app:details}.

\paragraph{Evaluation Datasets} We also need multilingual image captioning datasets for evaluation which are not observed during training. For this, we measure groundedness on three separate datasets, each with its own strengths and weaknesses. First, we use \textbf{Crossmodal-3600}. This dataset includes captions for 3,600 images across a range of cultures, manually  captioned by fluent speakers of 36 typologically diverse languages. However, it is relatively small per language compared to other datasets. Further, the independence of the captions means that there is greater diversity in what aspects of an image are being described across languages \citep{liu-etal-2021-visually,ye2024computervisiondatasetsmodels,berger2024crosslingualcrossculturalvariationimage}.

Our second dataset, the validation set of \textbf{COCO-35L}, addresses several of these issues. It is larger, with 5 captions each for 5000 images and 35 languages,\footnote{Crossmodal-3600 and COCO-35L cover the same languages with the exception of Quechua.} yielding 25,000 captions per language. Further, the captions are machine translations of each other, ensuring more comparable semantic content across languages \citep{beekhuizen2017} at the expense of centering the perspective of English speakers and machine translation issues.

Finally, we consider \textbf{Multi30K}. This dataset comprises 30,000 images captioned 5 times each in English, with a single caption per image manually translated into French, German, Czech, and Arabic. This dataset is therefore large on the individual language level, but with limited language coverage. It has the  comparability of being translated and the trustworthiness of human translation, but may still be vulnerable to translationese. By looking at all three of these datasets for similar generalizations about the relationship between groundedness and part of speech, we obtain a picture that is robust to the weaknesses of the individual datasets.

\section{Results}
\begin{figure}\label{fig:permutation}
  \centering
    \includegraphics[width=\linewidth]{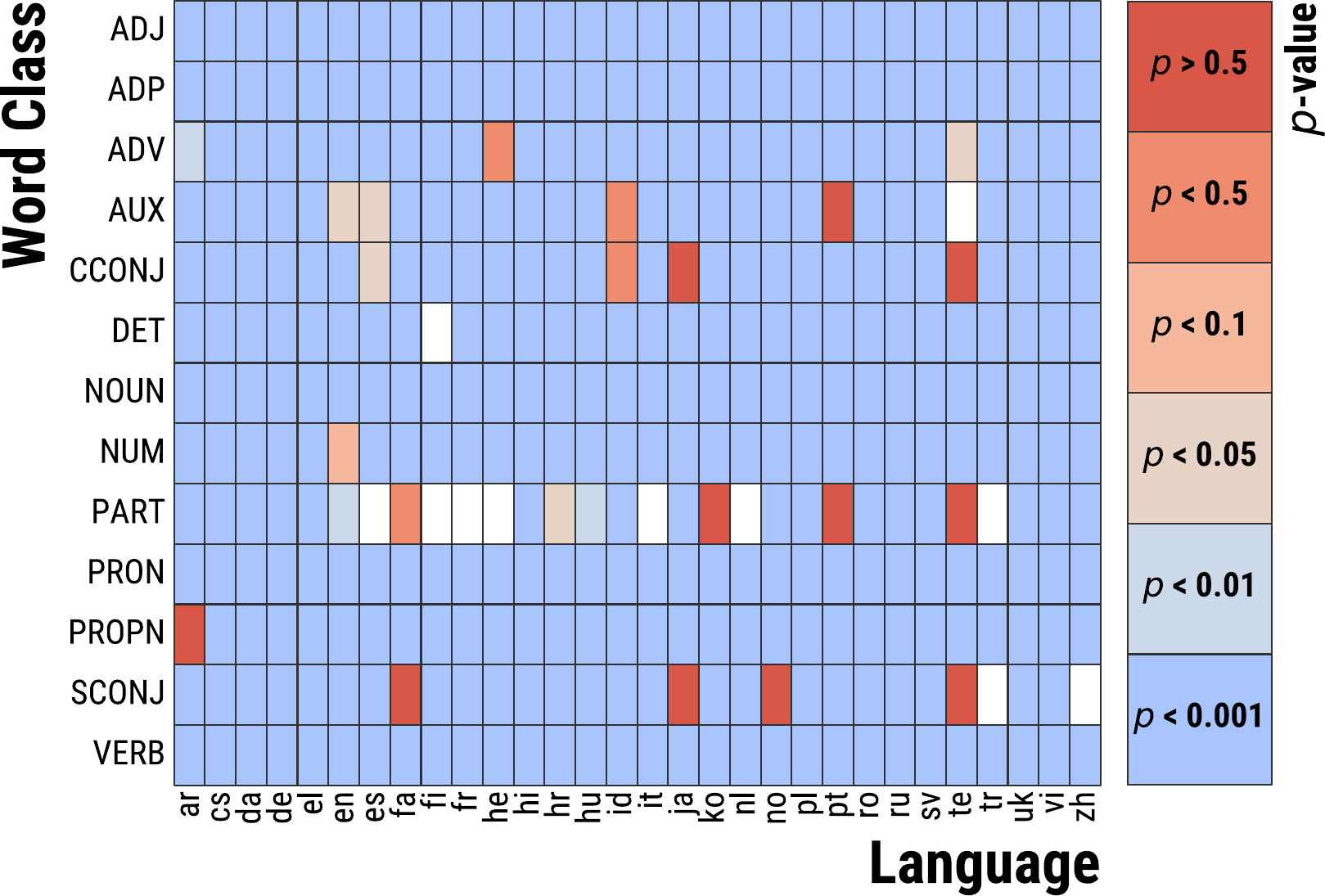}
  \caption{Heatmap of mutual information estimates across parts of speech in thirty languages. Cells show the statistical significance of a word class's groundedness (MI > 0). Unattested classes are white. Some functional classes display non-significant levels of groundedness in several languages, while lexical classes dominantly show highly significant grounding.}
  \label{fig:heatmap}
\end{figure}

The following sections quantitatively investigate the trends in our groundedness measure across languages and word classes. We begin by examining which word classes exhibit significant groundedness (Section~\ref{sec:perm}), followed by an analysis of cross-linguistic trends and their consistency (\ref{sec:order} and~\ref{sec:consistent}). Finally, we relate our findings to  contentfulness-related psycholinguistic norms (\ref{sec:human}).

\subsection{Which word classes are grounded?}\label{sec:perm}

We first investigate %
the evidence for groundedness in each word classs---that is, for each part of speech, we ask 
whether its estimated mutual information with the image is significantly greater than zero. 

To compute significance levels, we use a one-sample permutation test. Taking the set of PMIs for a part of speech (POS) in a language, we sample up to 500 PMIs at a time from all datasets and randomly permute their signs (assign + or - with equal probability to each PMI value), then average these values to produce a new estimate of mutual information (MI). We repeat this process to produce $10^5$ permuted estimates. By measuring how often our estimate based on the observed data is greater than the permuted estimate, we obtain the $p$-value,\footnote{We use the \citet{benjamini-control-2001} corrections.} i.e., the probability that our observations would have occurred under the null hypothesis of MI = 0.

Results are shown in Figure~\ref{fig:heatmap}. Overall, the results suggest most or all word classes contribute some information about the image they describe---in line with theories in linguistics that emphasize the lexical aspects of categories which are traditionally considered functional \cite{corver-semilexical-2001, bisang-grammaticalization-2017}. Interestingly, subordinating and coordinating conjunctions do not consistently reject the null hypothesis, suggesting there is little evidence the image is informative for how many clauses a speaker uses to describe an image.

\subsection{Which word classes are more grounded?}\label{sec:order}

\begin{figure*}[htbp]
    \centering
  \includegraphics[width=0.95\linewidth]{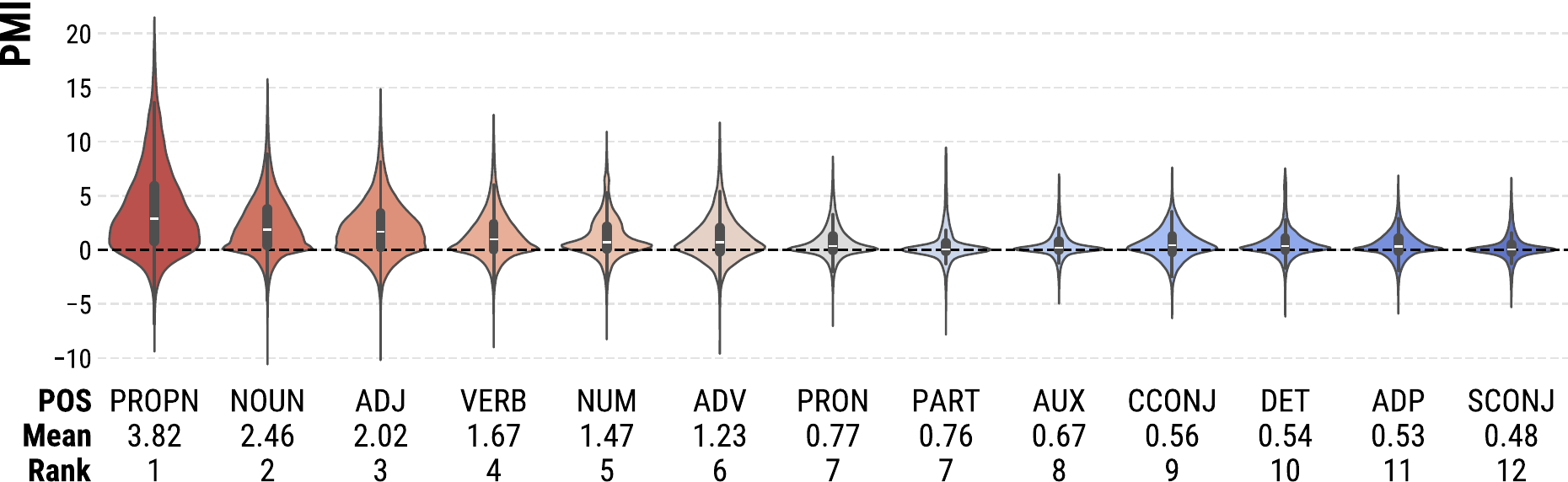}

    \caption{ Word token level distributions of the groundedness measure (PMI) across all languages and datasets, grouped by part of speech (word class). We also report the estimated marginal mean and ranking of each word class. Colors are based on the ranking of classes, rather than their average PMIs. Overall, the distribution and estimated ranking of word classes strongly suggest our groundedness measure quantitatively captures the distinction between lexical and functional classes.} %
    \label{fig:pos_ranking}
\end{figure*}
We hypothesize that the cross-linguistically consistent trends in word class groundedness correspond to a cline which is a continuous analogue of the lexical--functional word class distinction. To isolate the contribution of word class identity to mutual information cross-linguistically, we compute estimated marginal means (EMMs) for each word class's groundedness,\footnote{Averaged over values of language and dataset.} and perform a post-hoc pairwise comparison test of the means.\footnote{Using \v{S}id\'{a}k corrections; significance threshold $=0.01$.} The results of this analysis are displayed in Figure~\ref{fig:pos_ranking}. All pairwise comparisons except between pronouns and particles are statistically significant, leading to a near total ranking of word classes. We find that lexical word classes (Proper nouns, nouns, adjectives, verbs, numbers, and adverbs) have higher groundedness than functional word classes (particles, auxiliaries, conjunctions, determiners, and adpositions), with pronouns ranking together with particles at the upper end of the functional categories. %
The ranking corroborates ideas from cognitive linguistics which place nouns, adjectives, and verbs along a lexical--functional continuum, with nouns > adjectives > verbs \citep{rauhut-quantitative-2023}. %
On the other hand, it does not neatly align with ideas in linguistic theory about adpositions as a semi-lexical class \cite{corver-semilexical-2001}, which suggest they should behave more like other lexical classes compared to functional classes. Instead we see similar or greater mutual information for other functional classes, suggesting they could be more meaning-bearing than traditionally viewed.

\subsection{How consistent is word class groundedness across languages?}\label{sec:consistent}
We quantify the strength of the association between groundedness and word class on two levels: language-level MI estimates (Figure~\ref{fig:avg}), and token-level PMI ( Figure~\ref{fig:pos_ranking}). The first level quantifies how consistent languages are in the groundedness of word classes, while the second level quantifies how much word class drives the groundedness of individual tokens.  In both cases, we use ANOVA to estimate the amount of the variance in groundedness explained by word class.

\paragraph{MI estimates} For the language-level MI estimates in Figure~\ref{fig:avg}, we consider the separate effects of language, dataset, and POS on groundedness. Because the meanings (images) are matched across languages, this allows us to estimate and control for some languages having consistently larger or smaller MI estimates (due to language-specific variation in our neural estimators). We find significant effects of all 3 factors, but they differ dramatically in how much variation they explain. The effect of dataset is extremely small, explaining $0.5\%$ of the observed variance ($F_{3,816}\seq5.71$, $p \slt 0.01$). Language identity has a larger effect, explaining $8.2\%$ of the variance ($F_{29,789}\seq6.42$, $p\slt0.001$). However, word class dominates, explaining most of the total variance ($57.3\%$, $F_{12,806}\seq775$, $p\slt0.001$), and $62.8\%$ of the remaining variance after controlling for variance due to dataset and language. Altogether, these factors explain $65.6\%$ of the variance, leaving the remaining variance to cross-linguistic differences in the MI of specific parts of speech.

\begin{figure*}
\centering

\begin{subfigure}{.49\textwidth}
  \centering
  \includegraphics[width=0.9\linewidth]{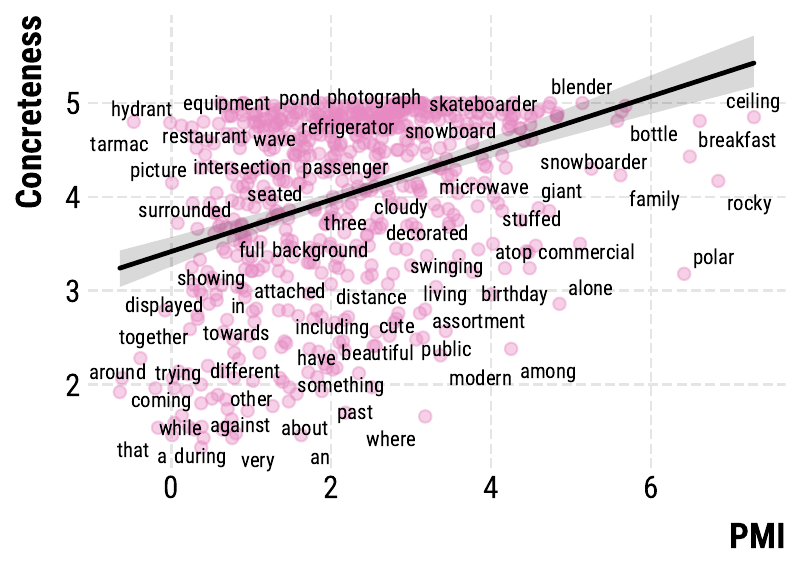}
  \label{fig:sub1}
\end{subfigure}
\begin{subfigure}{.49\textwidth}
  \centering
  \includegraphics[width=0.9\linewidth]{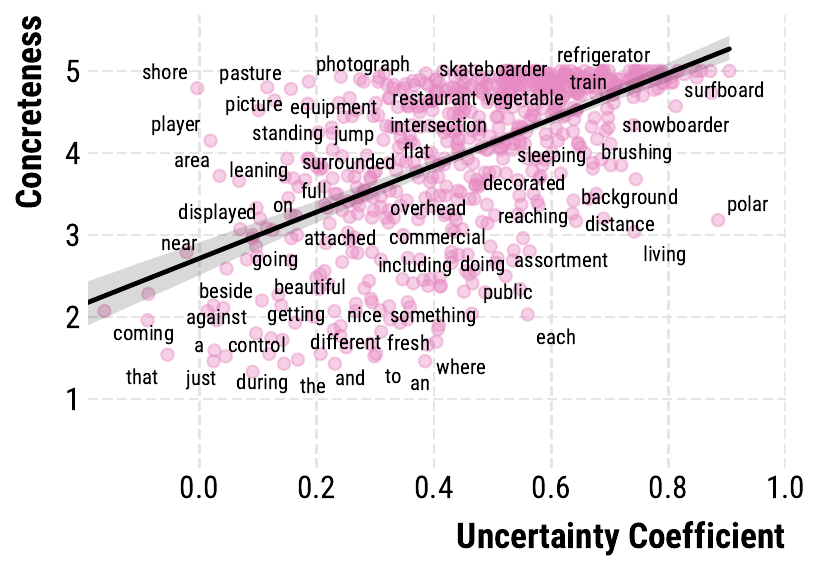}
  \label{fig:sub2}
\end{subfigure}
    \caption{Correlation between human concreteness ratings and type-level groundedness (PMI; left, $\rho\seq0.368$) or uncertainty coefficent (right, $\rho\seq0.609$): i.e., the average ratio between LM surprisal and captioning model surprisal.}
    \label{fig:norms}
\end{figure*}

\paragraph{PMI distributions} We also investigate how much variation in the full distribution of contextual groundedness estimates (PMIs) is explained by word class (shown in Figure~\ref{fig:pos_ranking}). Within a POS, groundedness is expected to vary substantially: for example, some (concrete, visually distinct) nouns have much higher PMI with the image than others, and tokens of the same word type also have different groundedness (e.g. ``lot'' referring to a location vs. ``lot'' as a quantity expression) Therefore, we expect word class to explain much less variance than in the overall MI estimates.
Language, dataset, and their interaction account for $2.4\%$ of the total variation in PMIs across the three datasets ($F_{64,10^7}\seq4727$, $p\slt0.001$). Word class accounts for $12.0\%$ of the total variation ($F_{12,10^7}\seq123583$, $p\slt0.001$). Additionally, the interaction between word class and language (cross-linguistic variation in the means of word classes) accounts for only an additional $1.6\%$ of the total variation ($F_{330,10^7}\seq602.5$, $p\slt0.001$), despite having many degrees of freedom. So cross-linguistically consistent tendencies comprise the bulk of the explainable variance in the overall PMI distribution across these three datasets---5 times as much as language and dataset, and 7.5 times as much as language differences in POS groundedness.%

\subsection{Semantic dimension of the measure}
\label{sec:human}
In this section we explore the semantic properties of the groundedness measure introduced here, comparing it to semantic norms related to contentfulness that are widely used in psycholinguistics. One potential advantage of our method is the ease with which it allows the rating of individual word tokens in context; however, existing ratings tend to be for words in isolation (word types). We focus our analysis here on English and on word types which occur at least 30 times in the COCO(-35L) validation set, %
averaging across occurrences to obtain an estimate of the average type-level groundedness. 

We compare to three different psycholinguistic norms: imageability, concreteness, and strength of visual experience. Such norms are measured by providing a definition and examples of low- and high-value words to raters, who then rate words on a Likert Scale. For imagability, we use the Glasgow Psycholinguistic Norms \citep{scott-glasgow-2019}. For concreteness, we use the\citet{brysbaert-concreteness-2014} norms. For strength of visual experience, we use the Lancaster Sensorimotor Norms \citep{lynott-lancaster-2020}. Results for concreteness are shown in Figure~\ref{fig:norms} (left). We observe fairly weak (though significant, $p\slt0.001$) correlations with groundedness using Spearman's $\rho$ (Imageability: $\rho\seq0.288$, Concreteness: $\rho\seq0.368$, Visual strength: $\rho\seq0.212$). 

We find these weak correlations are partly due to to the {\em informativity} aspect of our measures, which seems not to play as large of a role in human ratings (e.g. woman is just as concrete as skateboard, but less informative and also less grounded by our measure). To account for differences in baseline (LM) word informativity, we can normalize the PMI scores by the LM surprisal, yielding the uncertainty coefficient \citep{theil-estimation-1970}: the proportion of the LM surprisal explained by the PMI. Regressing this value against the psycholinguistic norms, stronger correlations emerge (Imagability: $\rho\seq0.548$, Concreteness: $\rho\seq0.609$ as shown in Figure~\ref{fig:norms} (right), Visual strength: $\rho\seq0.320$). This suggests that the differences between groundedness and surprisal are associated with concreteness. However, this measure collapses differences between word classes in overall informativity/surprisal.

In some cases, outliers are due to contextual effects. For example, 
in our data the word ``polar'' (high groundedness, moderate concreteness) occurs exclusively as the first word in the multiword expression ``polar bear'' which is highly concrete, imageable, and visual; while ratings based on the word type are for the more abstract geographical concept. Other words with divergent scores between human-based and model-based methods tend to be those which frequently occur in contexts where they are highly expected (e.g. ``shore'' which tends to occur in limited syntactic contexts and after the appearance of words like ``boat,'' ``lake,'' or ``surfers''), or words which are often used non-specifically in the image captioning context (e.g. ``photo'' exhibits very low PMIs, because  captions frequently begin with ``A photo of \dots'').

\section{Discussion and Conclusion}%
We have proposed a grounded approach to typology, using images as a proxy for sentence meaning. Using information theory and neural models, we define {\em groundedness}, a measure of a token's association with the meaning expressed in a sentence Our results demonstrate that word classes display consistent patterns in terms of their groundedness across a typologically diverse sample of languages. We find these patterns can be described as a continuous cline which generalizes the traditionally dichotomous distinction between lexical and functional word classes into a gradient one%
. However, our results suggest grammatical word classes still carry semantic content. We find that nouns > adjectives > verbs, in line with a view of these classes as a continuum; yet, our results contradict claims that adpositions are more lexical than other functional classes. Our measure is related to surprisal, but diverges from it, particularly for concrete words.

While this work has focused on word classes, groundedness enables the exploration of other aspects of how languages express function through form. %
Future work could investigate in detail under what conditions ``functional'' items have higher groundedness. For example, do more spatial adpositions and determiners have higher groundedness than less spatial ones?  Humans tend to have difficulty scoring highly abstract and grammaticalized words, and getting contextual scores is difficult with existing psycholinguistic  approaches: groundedness opens new ways to address these questions.

Our approach is also suitable for studying non-prototypical word class organizations, such as languages which do not clearly distinguish between adjectives and verbs (Korean; \citealp{maling1998case}), or languages that split individual word classes into distinct sub-classes (Japanese adjectives; \citealp{backhouse-1984}). Future work should look at both formal and semantic sub-classes of parts of speech---such as gerunds, participles, and different semantic classes of verbs (as in VerbNet; \citealp{kipper-schuler-etal-2009-verbnet})---investigating their groundedness and how it aligns with or varies from existing metrics. In particular, we conjecture that boundary classes (e.g. gerunds) may display intermediate groundedness (between nouns and verbs) compared to prototypical members of those classes. Groundedness makes it possible to test this conjecture with reference to the contexts in which words appear, which is needed for distinguishing syncretic forms.

Our approach can also cover any classes which can be defined over linguistic units, such as morphemes, phrases, or semantic classes. For instance, future work could explore the claim that inflections are more ``grammatical'' than derivations \citep{booij-inflection-2007, Haley_Ponti_Goldwater_2024}. Similarly, our measure could be used to study the lexicalization or grammaticalization of constructions (as a decrease in groundedness over time). To support such work, we release our groundedness scores online.\footnote{\url{https://osf.io/bdhna/}}

Going beyond the details of the approach here, our work generally suggests a role for multimodal models in computational typology similar to the one played by language models in the past decade \citep[e.g.][]{pimentel-etal-2023-revisiting,cotterell-etal-2018-languages,ackerman-morphological-2013}. While language coverage remains more limited than text models, the latest multimodal models and datasets cover enough typologically and culturally diverse languages to make them worth studying---and we anticipate coverage will only improve. Further, the ability of multimodal models to provide an empirically grounded (if imperfect) representation of meaning makes them uniquely valuable for quantitatively addressing questions about the relation between form and function in language. Our work provides the first study of this kind, and we hope that by demonstrating the utility of this approach and releasing our groundedness scores we will inspire other researchers to follow suit.

\section{Limitations}
Our approach has a number of important limitations. These limitations should inform the interpretation of results here, as well as any future studies considering using these techniques.

First, our operationalisation of meaning as an image is necessarily a simplification and has numerous implications for our results. Notably, the choice of images rather than videos (motivated by model quality and availability) as the representation of meaning has major implications for verbs, which tend to have meanings which are more temporally extended. This choice also has substantial implications about the variety of language which can be analyzed--many types of language use, such as metaphoric extension, are likely to be much less frequent in image captions than in other domains of language use: such phenomena are perhaps best studied using a different technique. This problem is compounded by the fact that existing multilingual corpora for these datasets remain fairly small--thus the analysis of long-tail phenomena in language using these methods is likely not yet possible.

Compared to existing methods in typology, this method trades human effort for computational resources. While we make both our models and data available, significantly lessening the burden on future studies, the models here contain between two and three billion parameters, and the image models have very long sequence lengths due to the image tokens. Inference on new data is therefore fairly expensive with current technologies.

Further, there remain significant limitations on the languages which can be studied with these approaches. Currently available models cover just 16 languages which are not part of the Indo-European language family, and entire areal typological regions like the Americas are not covered. We hope that the quality and coverage of these models can continue to improve, and that findings based on current models can be revisited and replicated with newer models.

Finally, we rely on automatic part of speech tagging based on Universal Dependencies for the analyses here (see Appendix~\ref{app:details} for further information). Overall, the accuracy of the Stanza tagger is high for the Universal Dependencies corpora of the languages studied here ($96\%$ on average); however, it is not uniformly accurate across languages. Vietnamese has the lowest average accuracy, with $81.5\%$ on their test set; however, our data is different in domain from many of the universal dependencies corpora, so the accuracy might be somewhat lower or higher. Universal Dependencies part of speech tags are not entirely without controversy as well---for instance, some linguists would argue that Korean does not have an adjective class, but UD uses one. It is possible that choices or inconsitencies in the assignment of POS tags according to UD could impact some MI estimates. In summary, noise due to POS tagging may have some influence on the results here, but is unlikely to affect our main conclusions.

\bibliography{anthology,custom,review}

\clearpage
\appendix
\onecolumn
\section{Implementation details}
\label{app:details}
\subsection{Part of Speech annotations}
Note that none of the datasets used here come annotated with word class information. We adopt the Universal Dependencies tagset, using Stanza \citep[v.1.8.2]{qi-stanza-2020} to tag words with their Universal Dependencies parts of speech. We remove single orthographic words that Stanza assigns multiple parts of speech, like English ``don't'' or German ``zum'' from our analysis, since it is unclear to which part of speech they should be assigned. Stanza does not cover Thai, Maori, Tagalog, Swahili, or Bengali for part of speech tagging, so they are excluded.
\subsection{Word-level PMI Estimates}
\label{app:word-prob}
Because the tokenizer of the present model does not cross orthographic word boundaries, we are able to sum the log probabilities of their constituent subword tokens to obtain word-level rather than token-level log probability estimates. Ordinarily, some languages do not indicate word boundaries in their orthography, such as Japanese; however, the pretraining data and evaluation datasets (Crossmodal-3600 and COCO-35L) are word-tokenized, so this information is readily available.
Further, because our language model uses sub-word tokenization with trailing whitespaces, we adopt the correction proposed by \citet{oh2024leadingwhitespaceslanguagemodels,pimentel2024computeprobabilityword}. Specifically, let $\mathbf{s}_{w_t}$ be the decomposition of word $w_t$ into a sequence of subwords, and $\mathbf{s}_{\mathbf{w}_{<t}}$ be the decomposition of context $\mathbf{w}_{<t}$ into a sequence of subwords.
Given $\mathcal{S}_{\text{bow}}$, the subset of the tokenizer vocabulary that contains subwords that are beginning-of-word (e.g., with a trailing whitespace):
\begin{align}
     p(w_t \mid \mathbf{w}_{<t}) = p(\mathbf{s}_{w_t} \mid \mathbf{s}_{\mathbf{w}_{<t}}) \cdot \frac{\sum_{s \in \mathcal{S}_{\text{bow}}} p(s \mid \mathbf{s}_{\mathbf{w}_{<t}} \odot \mathbf{s}_{w_t})}{\sum_{s \in \mathcal{S}_{\text{bow}}} p(s \mid \mathbf{s}_{\mathbf{w}_{<t}})}
 \end{align}
where $\odot$ stands for concatenation.
\subsection{Training details}
For training our language model, we did a grid search over learning rates and whether or not to use weight decay. We use a learning rate of $2\times10^{-5}$ and weight decay of $1\times10^{-6}$ with the Adam optimizer. To train the final model, we train on a single A100 with a batch size of 4 for 430,000 steps on COCO-35L ($\approx50$ hours of training, approximately 3 epochs). Our model achieves much lower perplexity on our evaluation datasets than Gemma-2B, suggesting successful domain adaptation.

\section{F-statistics}

\section{Correlation plots for other psycholinguistic norms}
\begin{figure*}
\centering

\begin{subfigure}{.49\textwidth}
  \centering
  \includegraphics[width=0.9\linewidth]{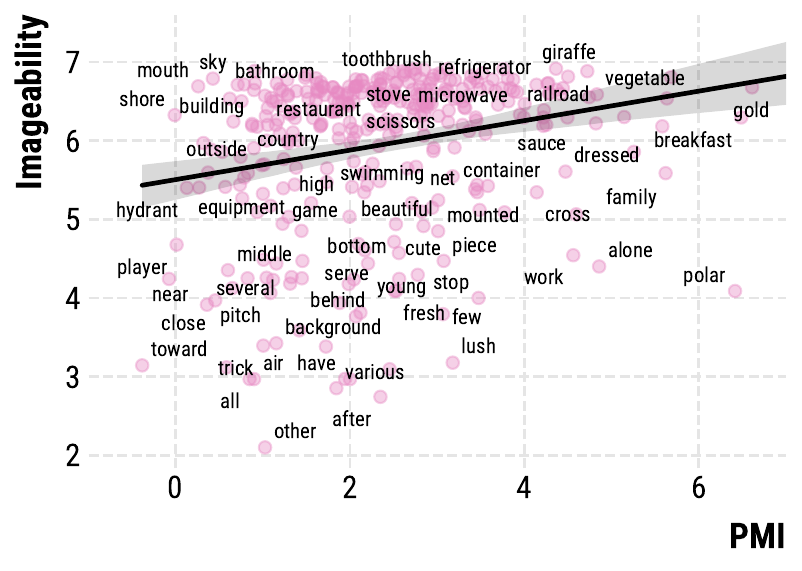}
  \caption{$\rho=0.288$}
  \label{fig:sub3}
\end{subfigure}
\begin{subfigure}{.49\textwidth}
  \centering
  \includegraphics[width=0.9\linewidth]{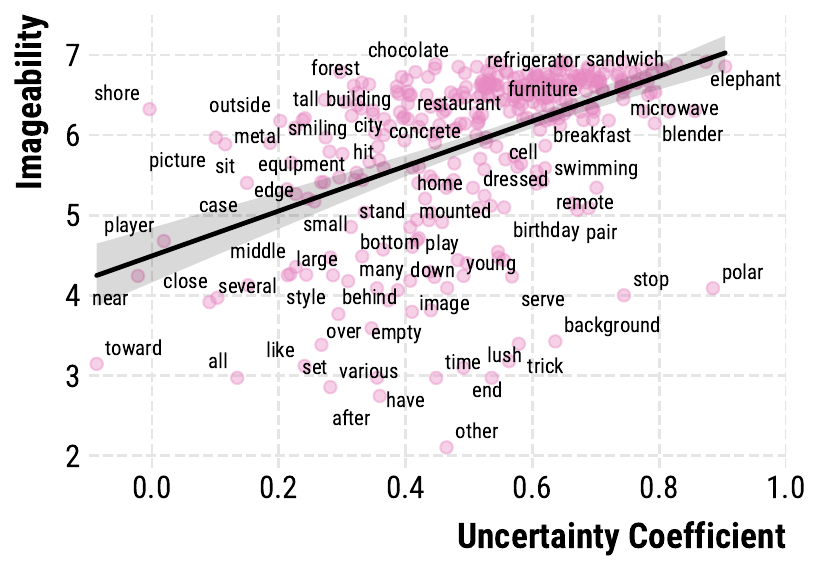}
\caption{$\rho=0.548$}
  \label{fig:sub4}
\end{subfigure}
\begin{subfigure}{.49\textwidth}
  \centering
  \includegraphics[width=0.9\linewidth]{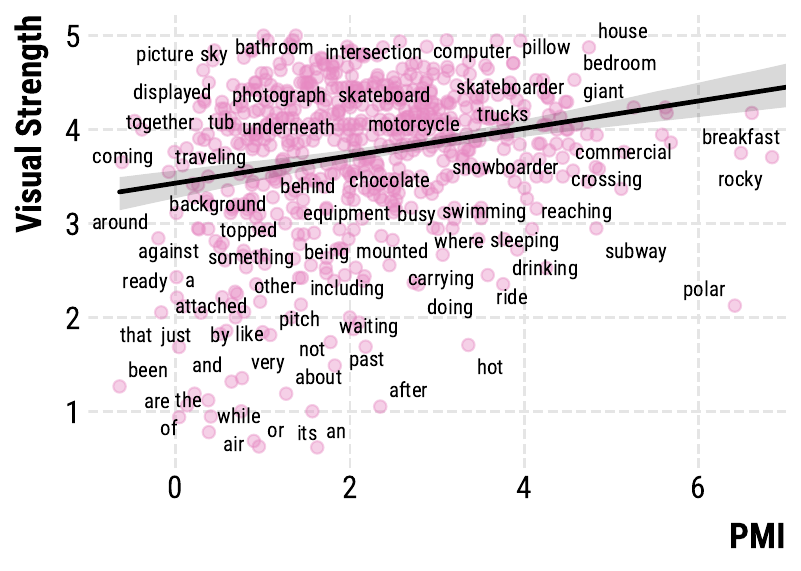}
  \caption{$\rho=0.212$}
  \label{fig:sub5}
\end{subfigure}
\begin{subfigure}{.49\textwidth}
  \centering
  \includegraphics[width=0.9\linewidth]{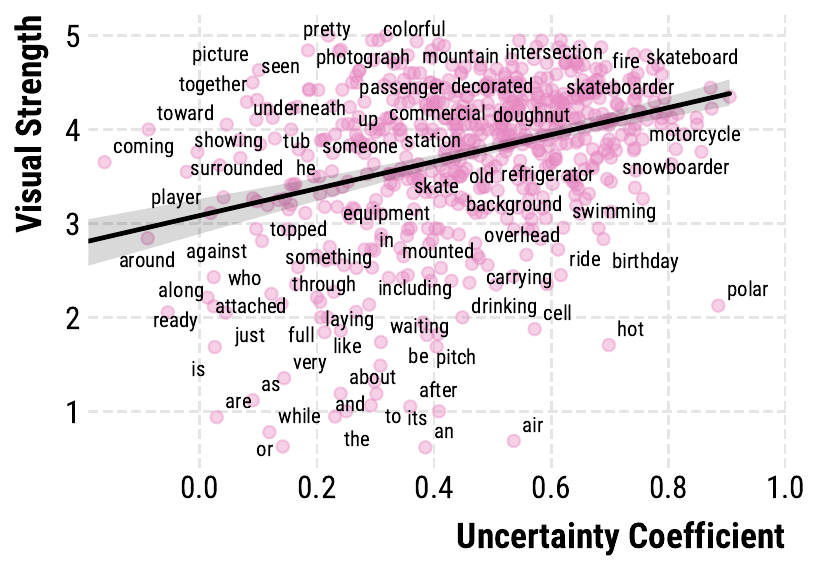}
    \caption{$\rho=0.320$}
  \label{fig:sub6}
\end{subfigure}
    \caption{Correlation between English psycholinguistic norms  and type-level groundedness  (left) or uncertainty coefficent (right): i.e., the average ratio between LM surprisal and captioning model surprisal. Type-level measures were computed by averaging scores across the COCO-dev dataset for types which occur at least 30 times.}
    \label{fig:norms_extra}
\end{figure*}

Figure~\ref{fig:norms_extra} shows the relationship between our measure and concreteness, as well as the uncertainty coefficient, which normalizes our measure by the language model surprisal. While concreteness is most strongly associated with our measure/its normalized variant, for completeness we show the relationships between our measure and the other psycholinguistic norms (imageability and strength of visual experience) we investigate here.
\twocolumn
\section{Groundedness distribution for Crossmodal-3600}\label{results:xm}
Results are ordered by descending mutual information estimate within the dataset (average groundedness/PMI). Hue indicates the average cross-linguistic ranking of a part of speech.

\vspace{1em}
\noindent
\foreach \langtwo in {ar,cs,da,de,el,en,es,fa,fi,fr,he,hi,hr,hu,id,it,ja,ko,nl,no,pl,pt,ro,ru,sv,te,tr,uk,vi,zh} {
        \includegraphics[width=\linewidth]{xm3600-2/pos_\langtwo.pdf}
}
\section{Groundedness distribution for Multi30K}\label{results:multi}
Results are ordered by descending mutual information estimate within the dataset (average groundedness/PMI). Hue indicates the average cross-linguistic ranking of a part of speech.
\vspace{1em}
\noindent
\foreach \langthree in {ar,cs,de,en,fr} {
        \includegraphics[width=\linewidth]{multi30k/pos_\langthree.pdf}
}

\section{Groundness distribution for COCO-35L Development Set}\label{results:coco}
Results are ordered by descending mutual information estimate within the dataset (average groundedness/PMI). Hue indicates the average cross-linguistic ranking of a part of speech.

\vspace{1em}
\noindent
\foreach \lang in {ar,cs,da,de,el,en,es,fa,fi,fr,he,hi,hr,hu,id,it,ja,ko,nl,no,pl,pt,ro,ru,sv,te,tr,uk,vi,zh} {
        \includegraphics[width=\linewidth]{coco35/pos_\lang.pdf}
}

\onecolumn

\end{document}